%% file: main.tex
\documentclass[acmsmall]{acmart}

\usepackage{balance} 
\usepackage{multirow}
\usepackage{multicol}
\usepackage{tikz}
\usepackage{amsmath,amsfonts,amsthm,mathrsfs}
\usepackage{diagbox}
\usepackage{caption}
\usepackage{subcaption}
\usepackage{graphicx}
\usepackage{booktabs}
\usepackage{stfloats}
\usepackage{float}
\usepackage{setspace}
\usepackage{svg}
\usepackage{url}
\usepackage{dsfont}
\usepackage{algpseudocode}
\usepackage{algorithm}
\usepackage{textcomp}
\usepackage{xcolor}
\usepackage{color, colortbl}
\usepackage{xspace}
\usepackage{pifont}
\usepackage{booktabs,siunitx}
\usepackage{longtable}
\usepackage{supertabular}
\usepackage{url}
\usepackage{setspace}
\usepackage[normalem]{ulem}
\usepackage{tabularx,tabulary}
\usepackage{makecell}
\usepackage{circledsteps}
\usepackage{tcolorbox}
\usepackage{lipsum} 
\usepackage{hyperref}
\usepackage[hyphenbreaks]{breakurl}
\usepackage{titlesec}

\usepackage{tcolorbox}
\usepackage{soul}

\usetikzlibrary{shadows}

\newtcolorbox{shadedbox}{
  drop shadow southeast,
  breakable,
  enhanced jigsaw,
  colback=lightgray
}

\definecolor{codegreen}{rgb}{0,0.6,0}
\definecolor{codegray}{rgb}{0.5,0.5,0.5}
\definecolor{codepurple}{rgb}{0.58,0,0.82}
\definecolor{backcolour}{rgb}{0.95,0.95,0.92}
\definecolor{keywordcolor}{rgb}{0.2,0.2,0.6}
\definecolor{amber}{rgb}{1.0, 0.75, 0.0}

\newcommand{\del}[1]{}

\newcounter{finding}
\AtBeginDocument{%
  }

\renewcommand\appendix{
  \par
  \setcounter{section}{0}
  \renewcommand\thesection{\Alph{section}}
  \renewcommand\thetable{\thesection\arabic{table}}
  \renewcommand\thefigure{\thesection\arabic{figure}}
  \renewcommand\theHsection{\thesection}
  \renewcommand\theHtable{\thetable}
  \renewcommand\theHfigure{\thefigure}
}

\begin{document}


\title{Enhancing Patient-Centric Communication: Leveraging LLMs to Simulate Patient Perspectives}


\author{Xinyao Ma}
\authornote{These authors contributed equally to this work.}
\orcid{0000-0002-4760-6072}
\affiliation{%
\institution{Indiana University Bloomington}
\city{Bloomington}
\state{IN}
\country{USA}}
\email{maxiny@iu.edu}

\author{Rui Zhu}
\authornotemark[1]
\orcid{0000-0002-8059-6718}
\affiliation{%
\institution{Yale University }
\city{New Haven}
\state{CT}
\country{USA}}
\email{rui.zhu.rz399@yale.edu}

\author{Zihao Wang}
\affiliation{%
\institution{Indiana University Bloomington}
\city{Bloomington}
\state{IN}
\country{USA}}
\email{zwa2@iu.edu}

\author{Jingwei Xiong}
\affiliation{%
\institution{University of California, Davis}
\city{Davis}
\state{CA}
\country{USA}}
\email{jwxxiong@ucdavis.edu}

\author{Qingyu Chen}
\affiliation{%
\institution{Yale University }
\city{New Haven}
\state{CT}
\country{USA}}
\email{qingyu.chen@yale.edu}

\author{Haixu Tang}
\affiliation{%
\institution{Indiana University Bloomington}
\city{Bloomington}
\state{IN}
\country{USA}}
\email{hatang@iu.edu}

\author{L. Jean Camp}
\orcid{0000-0001-8731-7884}
\affiliation{%
\institution{Indiana University Bloomington}
\city{Bloomington}
\state{IN}
\country{USA}}
\email{ljcamp@iu.edu}

\author{Lucila Ohno-Machado}
\affiliation{%
\institution{Yale University }
\city{New Haven}
\state{CT}
\country{USA}}
\email{lucila.ohno-machado@yale.edu}


\begin{abstract}

Large Language Models (LLMs) have demonstrated impressive capabilities in role-playing scenarios, particularly in simulating domain-specific experts using tailored prompts. This ability enables LLMs to adopt the persona of individuals with specific backgrounds, offering a cost-effective and efficient alternative to traditional, resource-intensive user studies. By mimicking human behavior, LLMs can anticipate responses and actions based on concrete demographic, professional, or experiential profiles. In this paper, we evaluate the effectiveness of LLMs in simulating individuals with diverse backgrounds and analyze the consistency of these simulated behaviors compared to real-world outcomes. In particular, we explore the potential of LLMs to interpret and respond to discharge summaries provided to patients leaving the Intensive Care Unit (ICU). We evaluate and compare with human responses the comprehensibility of discharge summaries among individuals with varying educational backgrounds, using this analysis to assess the strengths and limitations of LLM-driven simulations. Notably, when LLMs are primed with educational background information, they deliver accurate and actionable medical guidance 88\% of the time. However, when other information is included in the query, performance significantly drops, falling below random chance levels. This preliminary study sheds light on the potential benefits and pitfalls of automatically generating patient-specific health information for individuals from diverse backgrounds and populations. While LLMs show promise in simulating health personas, our results highlight critical gaps that must be addressed before they can be reliably used in clinical settings. Specifically, our findings suggest that a straightforward query-response model could outperform a more tailored approach in delivering health information. Introducing even simple patient-specific information can either improve or degrade the model’s effectiveness. We discuss these insights as an important first step in understanding how LLMs can be optimized to provide personalized health communication while maintaining accuracy and safety.


\end{abstract}

\keywords{Large Language Models, ChatGPT, Discharge Summary, Role-playing}

\maketitle

\input{1.introduction}

\input{2.background}
\input{3.method}
\input{4.results}

\input{5.discussion}
\input{6.relatedwork}

\input{7.conclusion}


\bibliographystyle{ACM-Reference-Format}
\bibliography{reference}

\input{appendix}

\end{document}

%% file: 1.introduction.tex
\section{Introduction}

Large Language Models (LLMs) have revolutionized natural language processing (NLP) and artificial intelligence (AI) across a range of tasks, from text generation to complex reasoning. A key aspect of their versatility is their capability for in-context learning (ICL), which allows LLMs to dynamically adapt to different roles and instructions without requiring retraining~\cite{GPT3}. This flexibility makes LLMs highly suitable for role-playing tasks, where they can be quickly configured to simulate individuals with specific demographic, professional, or experiential attributes. Through ICL, LLMs can be tailored to serve as agents for various applications, offering an efficient and adaptable alternative to traditional user studies.  

Recent studies have highlighted the potential of LLMs in role-playing scenarios. Zero-shot reasoning abilities have been leveraged with role-play prompting~\cite{DBLP:conf/naacl/KongZCLQSZWD24} to explore demographic biases by emulating age-specific behaviors~\cite{DBLP:conf/nips/SalewskiARSA23}. In addition, LLMs have been applied as both subjective and objective evaluators in tasks like text summarization~\cite{DBLP:conf/nlpcc/WuGSLJ23} and qualitative coding of health-related free-text survey data~\cite{lossio2024comparison}, illustrating the breadth of perspectives these models can provide with targeted queries. Beyond individual tasks, frameworks such as the Role-Playing Framework~\cite{DBLP:conf/nips/LiHIKG23} allow LLMs to collaborate in multi-agent environments, opening new possibilities for cooperative problem-solving.

In this study, we focus on the ability of LLMs to role-play individuals from diverse backgrounds, with a particular emphasis on simulating the comprehension of discharge summaries in clinical settings. Discharge summaries often contain complex medical information that can be challenging for patients with limited medical knowledge to fully grasp. Therefore, tailoring these summaries to align with a patient’s background—such as their level of education, frequency of doctor visits, and other demographic factors—is crucial for effective communication and informed decision-making in healthcare. 

To implement this approach, we crafted role-specific prompts to guide the LLM’s behavior. 
For each scenario, we began with the statement \textit{``If you were a \{persona\}''}, where ``persona'' is replaced by a description of the target identity, such as educational attainment, socioeconomic status, or medical experience. This structured prompting allows the LLM to adapt its responses based on the specified profile, enabling it to mimic individuals with 
distinct levels of medical literacy, ranging from healthcare professionals to patients with no prior medical training. 
Our assessment focuses on both the consistency and accuracy of the LLM’s ability to understand and convey complex medical content to these different audiences. By comparing the outcomes of these LLM-driven simulations with real-world behaviors and reactions, we seek to determine how accurately these models replicate the nuances of human understanding and identify any limitations in their ability to adapt to diverse populations.

This work also explores the limitations of LLMs in role-playing such nuanced interactions, revealing challenges such as over-simplification, potential biases, and the boundaries of LLM comprehension. Through this analysis, we identify key areas where current models may fall short in conveying patient-specific information while recognizing their potential as a cost-effective and scalable solution for personalized communication. 

Ultimately, this research represents a first step toward leveraging LLMs to automatically produce tailored discharge summaries that are more accessible and understandable for individuals from varied backgrounds. Yet our results showed that the current LLMs are failing for traditionally underserved populations. By addressing the communication gap between medical professionals and patients, 
especially those from underserved or linguistically diverse populations; this approach holds promise for improving patient outcomes and fostering more equitable healthcare communication. 


%% file: 2.background.tex
\section{Background}

\subsection{Large Language Models}
Large Language Models (LLMs)~\cite{GPT3, GPT-4, PaLM, PaLM2, LaMDA} have demonstrated exceptional capabilities in creatively addressing diverse NLP and AI tasks. Recent studies have utilized instruction tuning\cite{Instruction-Finetuned, Multitask_Prompted, Finetuned_LM} and reinforcement learning from human feedback (RLHF)~\cite{human_feedback, Harmless_RL} that are better aligned with human understanding and develop advanced AI assistants. Efforts are also underway to construct open-source LLMs~\cite{LLaMA, OPT, BLOOM, Pythia} to accelerate both research and industrial advancements. 

A significant aspect of LLMs' success is their in-context learning (ICL) capabilities~\cite{GPT3}, which enables models to carry out tasks based on the context provided in the input without requiring explicit retraining. This dynamic adaptability allows LLMs to respond to a wide range of prompts and conditions with increased flexibility. 
By leveraging ICL, models can assume different roles and perspectives, enhancing their reasoning capability and responsiveness through role playing. The growing emphasis on ICL is crucial for broadening the scope of LLM applications, especially in tasks that demand nuanced adaptation, such as collaborative problem-solving and personalized AI assistance.

\subsection{Discharge Summary and LLM-based Summary Generation}
Medical discharge summaries serve as vital communication tools between patients and healthcare providers, ensuring continuity of care after patients leave the hospital. These summaries typically include essential information such as medications, diagnoses, prescriptions, and follow-up instructions and often contain professional medical terms or abbreviations. It is crucial for patients to understand this information in order to adhere to their treatment plans and manage their recovery. Studies have shown that a high-quality discharge summary should be concise, delivered promptly, and contain relevant data. In a survey of 100 hospital-based physicians, respondents emphasized the importance of focusing on pertinent information that patients can easily comprehend and apply to their post-discharge care~\cite{van1999necessary}.

As a result, there has been growing interest in leveraging LLMs to automatically generate or assist medical discharge summaries~\cite{waheeb2020machine, chintalapati2023textual}. LLMs, with their advanced natural language processing capabilities, offer the potential to generate summaries that are not only accurate but also customized to the patient’s comprehension level. LLM-based summary generation is showing great potential in healthcare. By analyzing large sets of medical data and fine-tuning these models with domain-specific knowledge~\cite{kumichev2024medsyn}, and a tool, such as ``Discharge Me!’’ can automatically generate discharge summaries of clinical electronic health record~\cite{wu2024epfl}. Unlike our study, they did not evaluate medical accuracy or alignment with human subjects. The goal of these discharge summary generation tools is to capture the most important medical information while keeping things simple and clear.



\subsection{LLMs Personas}
As the ICL capabilities of large models continue to be uncovered, a growing body of work has focused on in-context role-playing and assisted evaluations. While human evaluation remains an option, manually assessing generated text is challenging, resource-intensive, and difficult to scale or repeat as requirements and quality criteria evolve. Consequently, many researchers have begun utilizing LLMs as natural language evaluators, demonstrating promising results~\cite{desmond2024evalullm, kim2024understanding}. Kong et al.\cite{DBLP:conf/naacl/KongZCLQSZWD24} proposed enhancing the zero-shot reasoning abilities of LLMs through role-play prompting, while Salewski et al.\cite{DBLP:conf/nips/SalewskiARSA23} explored the proficiency and biases of LLMs in emulating age-dependent behaviors. Role-playing has also been applied to a wide range of tasks. Wu et al.\cite{DBLP:conf/nlpcc/WuGSLJ23} employed LLMs as both objective and subjective role-players to evaluate summary texts, creating a more comprehensive evaluation framework by integrating insights from multiple role-based perspectives. Moreover, Li et al.\cite{DBLP:conf/nips/LiHIKG23} introduced a Role-Playing Framework, where agents assume various roles within a cooperative intelligence framework. This enables agents to autonomously collaborate to achieve tasks, offering new avenues for exploring cooperative strategies and synergies in multi-agent environments.

%% file: 3.method.tex
\section{Impersonation Methodology}
Our method consists of two main steps. First, we prompt the LLM using specific role-playing instructions. Second, we evaluate how closely the LLM's generated responses align with those of real human individuals in similar contexts.

\subsection{Prompting Large Language Model with Personas}
LLMs are trained to predict the most probable next token, \( t_k \), based on the sequence of preceding tokens, \( t_1, \dots, t_{k-1} \), by maximizing the likelihood function \( p_{\text{LLM}}(t_k|t_1, \dots, t_{k-1}) \). This approach enables the model to learn linguistic patterns and generate coherent responses. In this study, we employ pre-trained LLMs without any further fine-tuning, relying entirely on their existing knowledge and capacity to understand and generate natural language.

For each task, we generate one or more tokens by providing a task-specific context \( \boldsymbol{c} \), which is designed to instruct the LLM on how to approach the problem. The context typically includes a description of the task along with a specific prompt that guides the model towards a relevant answer. In particular, for impersonation tasks, we prefix the context with the instruction \textit{``If you were a \{persona\}''}, where \( p \) represents the persona to be simulated. This could be a social identity, such as ``a patient with no medical training'', or an area of expertise, such as ``a healthcare professional''.
By doing so, we tailor the LLM's behavior to fit the expected perspective of the impersonated persona, encouraging the model to generate responses as if it were that individual. This process leverages the model's in-context learning capabilities, allowing it to assume different roles based solely on the information provided in the prompt, without requiring explicit retraining.

Once the context is defined, we sample the generated tokens \( \boldsymbol{t} \) from the following distribution:
\[
p_{\text{LLM}}(\boldsymbol{t}|\boldsymbol{c}^{(p)}) = \prod_{k=1}^K p_{\text{LLM}}(t_k|c_1^{(p)}, \dots, c_n^{(p)}, t_1, \dots, t_{k-1})
\]
where \( c_1^{(p)}, \dots, c_n^{(p)} \) represent the context tokens related to the specific persona \( p \), and \( t_1, \dots, t_{k-1} \) represent the previously generated tokens. This probabilistic approach allows the model to iteratively generate a sequence of tokens based on both the task context and the previous outputs, ensuring continuity and coherence in the response.

We term this approach of shaping the model’s behavior based on the provided context as “in-context impersonation.” 
Using this method, the LLM dynamically adjusts its responses to reflect the persona’s perspective, demonstrating its ability to role-play different individuals based on minimal input cues. This technique is particularly useful for generating tailored outputs in complex scenarios, such as simulating how individuals with varying levels of medical knowledge might interpret healthcare information~\cite{kumichev2024medsyn, wu2024epfl, van1999necessary}.

In this paper, {\em in-context impersonation} forms the foundation of our evaluation framework, allowing us to assess how well LLMs can simulate and align with real human behaviors and perspectives across diverse social and professional contexts. This method highlights the flexibility of LLMs in adapting to different personas and underscores the importance of context-driven prompting in leveraging LLMs for role-specific tasks.

\subsection{Human Subjects Discharge Summary Evaluation}

Discharge summaries serve multiple purposes, with one of their key functions being to facilitate communication between clinicians and patients. A high-quality discharge summary should accurately convey medical advice while being easy for patients to understand. To assess patient comprehension and compare their responses to those generated by current LLMs, we conducted a human subject survey focusing on the evaluation of a given discharge summary. Participants read a section of a discharge summary and then answered ten follow-up questions aimed at assessing their understanding and recall of the information provided.

The questions were divided into two categories: information-based and perception-based. The eight information-based questions tested whether participants could accurately extract information from the discharge summary. These included direct questions such as “Do you know the name of all your medications?” and “Do you know your diagnosis?” as well as questions that need context understanding like “Do you know what kind of treatment you need to follow based on the discharge instructions?” These questions are intended to evaluate the human subjects' ability to acquire correct information from the discharge instructions.

The perception-based questions assessed participants' perceived difficulty in understanding the summary. The first perception question, ``Q1'' in Appendix~\ref{apdx:prcp_q}, was placed first, immediately after reading the discharge instructions, asking participants to rate their understanding level with options ranging from ``Very clear'' to ``Somewhat clear'' and ``Not clear at all.'' A second perception question was included at the end of the questionnaire for cross-validation, asking participants to rate the difficulty of understanding the discharge instructions, with responses ranging from "Extremely easy" to "Extremely difficult." Details of left experiment questions and summary are provided in Appendix~\ref{apdx:survey_questions}.

%% file: 4.results.tex
\section{Results}

\subsection{Data and Setup}



\noindent\textbf{Human Response Collection}.
All recruitment and enrollment procedures were reviewed and approved by the institutional IRB. We recruited participants from two groups based on their educational achievements (measured by degree received), as recorded by the Prolific platform. In addition, we included our own demographic questions covering self-reported education level, race, doctor visits, and emergency room visit frequency. The detailed demographic questions can be found in Appendix~\ref{apdx:survey_questions}. Each participant was required to read a section of the provided discharge summary and answer ten follow-up questions. They were allowed to refer back to the discharge summary while answering. To account for potential order effects and reduce reading fatigue, we randomized the presentation of three different discharge summaries across participants. 

\input{demo_tables}

Participants were compensated \$3 for completing the survey, based on an estimated completion time of 10 to 15 minutes. A \$1 bonus was offered for high-quality responses, defined by thoughtful engagement, such as taking more than the minimum possible time to complete the survey and answering all questions. In total, we received 96 valid responses from participants with diverse educational backgrounds. Among them, 49 identified as having a college or post-graduate degree, 47 had not received a college degree, and 35 reported having only a high school diploma. The demographic information for these participants is provided in Table~\ref{tab:demographics}.

\begin{figure}[ht]
    \centering
    \includegraphics[width=.65\columnwidth]{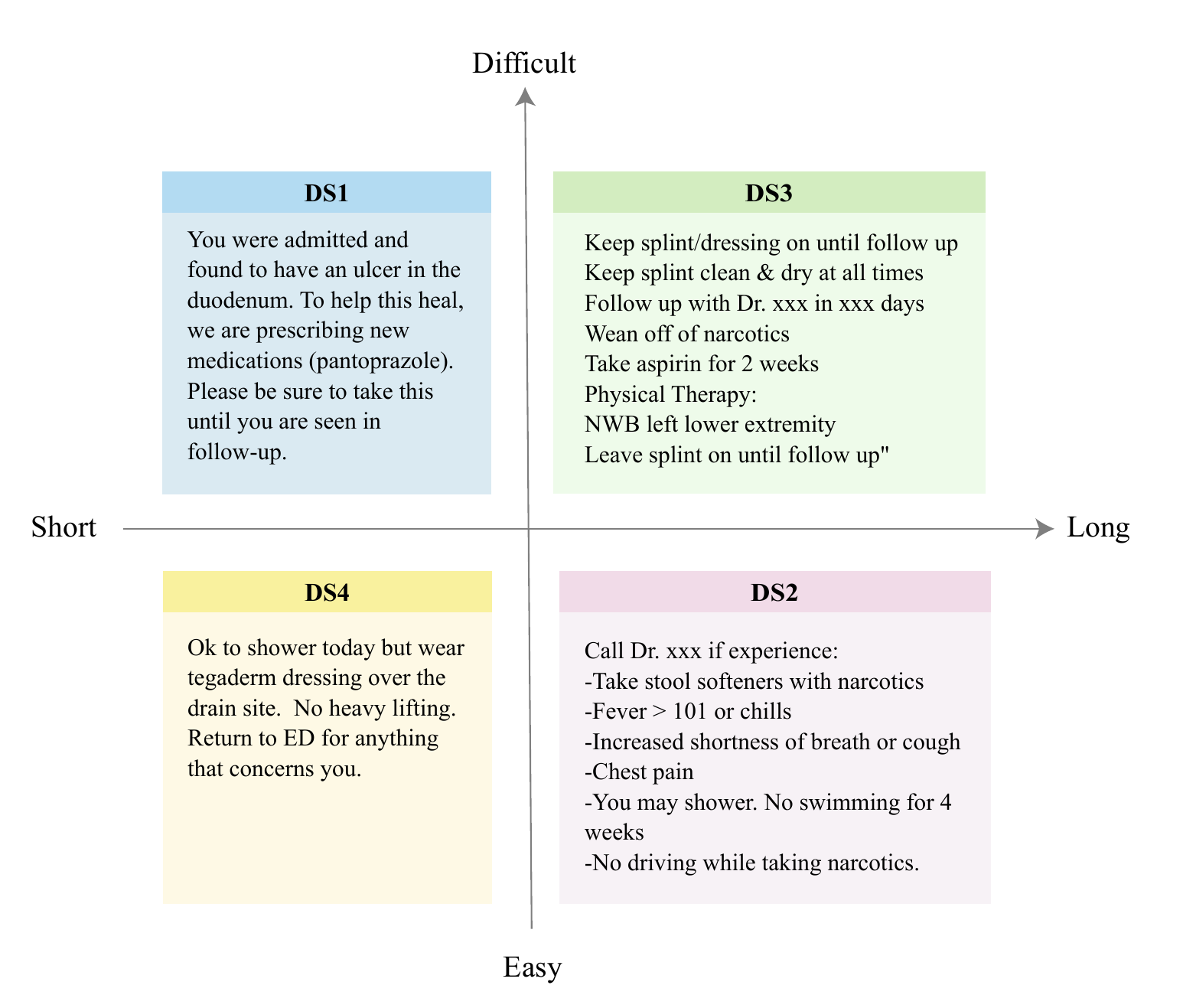}
    \vspace{-.1in}
    \Description[<Discharge Summary Selection:>]{<Four discharge summaries on different categories.>}
   \caption{Four discharge summaries on different categories.}
    \label{fig:discharge}
\end{figure}

\noindent\textbf{Discharge Summary Selection}. The discharge summaries were selected from an anonymized medical database, specifically focusing on sections intended for patient communication. Our selections balanced length and the difficulties in understanding, for example, if it contains professional medical words or abbreviations. We select one sample from each Discharge Summary (DS) category: DS1: short and hard to understand, DS2: long and easy to understand, DS3: long and hard to understand, and DS4: short and easy to understand. These sections included details on medications, treatments, and recommendations relevant to the patient's condition. Details of the discharge summary contents and categories explanations can be found in \autoref{fig:discharge}.

\noindent\textbf{Personas Considered}.
The first key question we explored was whether LLMs could effectively simulate the behaviors of individuals from different groups based on: education level, gender, frequency of doctor
visits, and frequency of emergency room (ER) visits. To investigate this, we prompted the LLM to imagine itself as a patient given certain background information, like ``you are someone who has never received a college degree.'' or ``you are someone whose gender identified as female''. The following is a description:``You will be reading a Discharge Summary written by a clinician for a patient. After reading the Discharge Summary, I will ask you a question. Your job is to think of someone with your background and choose the correct answer by selecting the letter of the answer (e.g., A, B, C, etc.). You should only output the letter with your final answer without any explanation.'' A prompt example can be found in Appendix~\ref{apdx:propt}.

\noindent\textbf{Large Language Models Considered}. 
We utilize the OpenAI API for GPT, specifically employing the \texttt{GPT4\_0613} model provided by Azure OpenAI API service. 


\subsection{Effectiveness}
\begin{table}[!t]
\setlength{\tabcolsep}{0.5mm}{
\footnotesize
\caption{Performance Summary. The similarity of the human response and the response of the corresponding persona for eight information-based questions and two perceptions questions for four discharge summaries.}
\label{tab:main}
\begin{tabular}{l|c|c|c|c|c|c|c|c|c}
\toprule
 & Low Edu. & High Edu. & Male & Female & Low Dr\_visit & High Dr\_visit & Low ER\_visit& High ER\_visit & Random Guess\\ 
\midrule
Infor. & \textbf{0.72}           & \textbf{0.88}          & \textbf{0.97}        & 0.44           & 0.47                     & 0.44                      & 0.31                     & 0.44          &     0.278        \\

Percep.  & \textbf{0.75}           & \textbf{0.75}            & 0.25         & 0.375          & 0.625                  & 0.5                       & 0.375                    & 0.5       &    0.267            \\
\bottomrule
\end{tabular}
}
\end{table}

In this section, we evaluate the overall effectiveness of the LLM's ability to simulate personas. We conduct experiments involving four distinct human group attributes: education level (low/high), gender (male/female), frequency of doctor visits (low/high), and frequency of emergency room (ER) visits (low/high). The experimental setup includes two types of tasks: information-based and perception-based. We also provided the random guess similarity accuracy based on the overall accuracy of the questions if they were answered on a strictly random basis. Specific statistics are presented in Table~\ref{tab:main}.

Our results show that the LLM achieves a significantly higher alignment rate compared to random guessing, suggesting that it has the ability to simulate human-like personas for some tasks and groups. Specifically, the average alignment rate across all tasks and human groups is 54.97\%, whereas random guessing would only achieve an alignment rate of approximately 26.7\%. This highlights the LLM's potential to model human-like responses and adapt to different personas effectively.

\begin{tcolorbox}[left=1mm, right=1mm, top=0.5mm, bottom=0.5mm, arc=1mm]
\textbf{Finding 1:}  
The LLM demonstrates strong persona capabilities in information-based tasks but struggles with perception-based tasks, particularly when simulating human understanding of medical discharge information.
\end{tcolorbox}

A deeper look at the task types reveals that the LLM performs better on information-based tasks, achieving an average alignment rate of 58.38\%, compared to perception-based tasks, where the average alignment rate is only 51.56\%. This performance gap may be attributed to the fact that individuals with similar attributes tend to show more consistent and predictable cognitive patterns when handling factual information, yielding homophilous results. This makes these tasks easier for the LLM to model. In contrast, perception-based tasks involve more diverse interpretations and personal preferences, making it harder for the LLM to simulate these responses.

\begin{tcolorbox}[left=1mm, right=1mm, top=0.5mm, bottom=0.5mm, arc=1mm]
\textbf{Finding 2:}  
The LLM excels at simulating individuals based on their educational background but struggles when simulating those with varying frequencies of emergency room (ER) visits.
\end{tcolorbox}

Among the four different human group attributes, the LLM performs best when simulating personas based on educational background. It achieves an average alignment rate of 77.50\% when modeling individuals with specific education levels, compared to a lower alignment rate of 40.63\% for individuals with different frequencies of ER visits. This discrepancy may be due to the fact that individuals with similar education backgrounds tend to have more clustered and consistent cognitive and behavioral patterns, making them easier to predict. In contrast, the group with varying ER visit frequencies is more diverse, reflecting broader variations in health conditions, behaviors, and personal experiences, which makes them more difficult for the LLM to simulate accurately.

Within the education category, the LLM shows a higher persona capability in simulating individuals with higher education levels. For example, when modeling the actions of individuals with a higher education background, the LLM achieves an average alignment rate of 85.4\%, compared to 72.6\% for individuals with lower education levels. This may be because higher education often correlates with more centralized, consistent patterns of behavior and thought processes, making it easier for the LLM to generate aligned responses. However, it is important to note that high persona capability in simulating individuals with higher education does not necessarily translate into equally strong performance when simulating individuals with lower education. These two groups do not exhibit directly opposing behaviors but rather reflect differences that the LLM may need to account for through more nuanced adjustments in its model.


\begin{figure}[ht]
    \centering
    \begin{subfigure}[t]{.35\columnwidth}
        \centering
        \includegraphics[width=\textwidth]{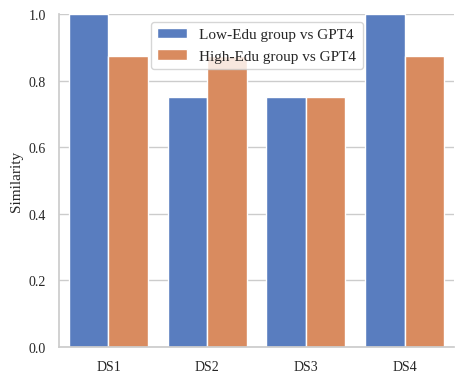}
        \caption{Information-Based Questions}
        \label{fig:infor_bar}
    \end{subfigure}
    \begin{subfigure}[t]{.35\columnwidth}
        \centering
        \includegraphics[width=\textwidth]{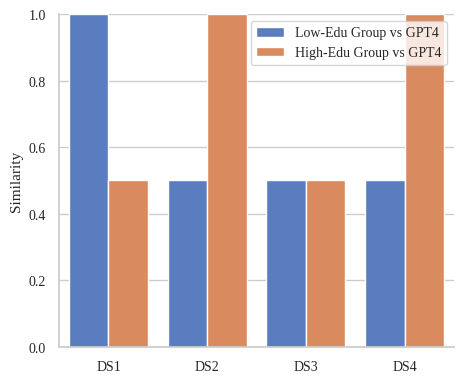}
        \caption{Perception-Based Questions}
        \label{fig:prcp_bar}
    \end{subfigure}
\caption{Effectiveness of the LLM's Ability to Simulate Personas Across Different Discharge Summary Categories}
\vspace{-.1in}
\label{fig:edu_bar}
\end{figure}

\subsection{Impact of Categories in Discharge Summaries}
Recall that we divided the discharge summaries (DS) into four categories: DS1 (short and hard to understand), DS2 (long and easy to understand), DS3 (long and hard to understand), and DS4 (short and easy to understand). In this section, we analyze how the DS category impacts the LLM's persona-simulation capability, using education level as a case study.

\begin{tcolorbox}[left=1mm, right=1mm, top=0.5mm, bottom=0.5mm, arc=1mm]
\textbf{Finding 3:} The alignment rate for simulating individuals with lower education levels is mainly driven by short discharge summaries.
\end{tcolorbox}

As shown in \autoref{fig:infor_bar}, the LLM achieves its highest alignment rates when simulating individuals with low education levels for the DS1 and DS4 categories, both of which feature short summaries. This suggests that individuals with lower education levels respond more consistently to shorter discharge summaries, likely because these summaries reduce cognitive load and are less demanding to comprehend. Given that individuals with lower education levels may have limited reading proficiency, shorter texts could allow for more straightforward processing, resulting in a higher alignment between the LLM’s simulation and actual human responses.

Additionally, we see that the LLM struggles to accurately simulate low-education personas with longer, more complex summaries (DS3). This may be due to the fact that longer or harder summaries introduce additional layers of complexity, such as advanced vocabulary or medical jargon, making it more difficult for both the LLM and individuals with lower education levels to process and respond in a predictable manner.

\begin{tcolorbox}[left=1mm, right=1mm, top=0.5mm, bottom=0.5mm, arc=1mm]
\textbf{Finding 4:} The alignment rate for simulating individuals with higher education levels is mainly driven by easier-to-understand discharge summaries.
\end{tcolorbox}

From \autoref{fig:prcp_bar}, we observe that the LLM achieves its highest alignment rates for individuals with higher education levels when handling the DS2 and DS4 categories, both of which are characterized as easy to understand. This suggests that individuals with higher education levels respond more consistently to discharge summaries that are clear and concise, regardless of length. Their stronger reading comprehension skills may enable them to process both short and long summaries effectively, as long as the content is presented in an accessible and straightforward manner.

Conversely, when the LLM simulates responses for more complex discharge summaries (DS1 and DS3), the alignment rate drops, even for individuals with higher education levels. This finding suggests that the difficulty of the content itself plays a significant role in reducing response predictability. Both high- and low-education groups tend to exhibit more diverse and unpredictable behaviors when faced with difficult-to-understand summaries, likely due to the increased cognitive effort required to interpret medical jargon or ambiguous phrasing.

Thus, the LLM's ability to simulate human responses is not solely influenced by education level but is also affected by the complexity of the information presented. While higher education groups tend to perform well with easy-to-understand texts, more difficult summaries can challenge even the most educated individuals, leading to more varied and less predictable responses.

\begin{figure}[ht]
    \centering
    \begin{subfigure}[t]{.35\columnwidth}
        \centering
        \includegraphics[width=\textwidth]{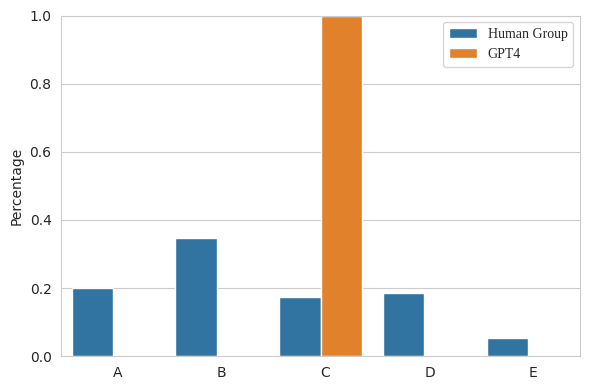}
        \caption{Answer distribution of DS1 Q10.}
        \label{fig:ds1_q10}
    \end{subfigure}
    \begin{subfigure}[t]{.35\columnwidth}
        \centering
        \includegraphics[width=\textwidth]{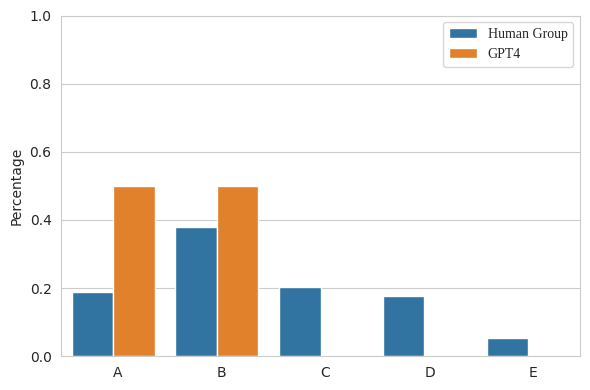}
        \caption{Answer distribution of DS2 Q10.}
        \label{fig:ds2_q10}
    \end{subfigure}
        \begin{subfigure}[t]{.35\columnwidth}
        \centering
        \includegraphics[width=\textwidth]{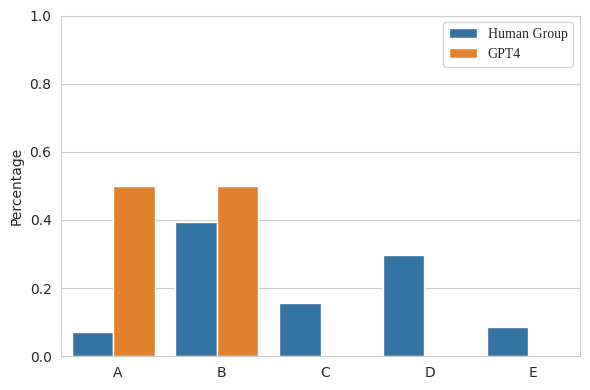}
        \caption{Answer distribution of DS3 Q10.}
        \label{fig:ds3_q10}
    \end{subfigure}
        \begin{subfigure}[t]{.35\columnwidth}
        \centering
        \includegraphics[width=\textwidth]{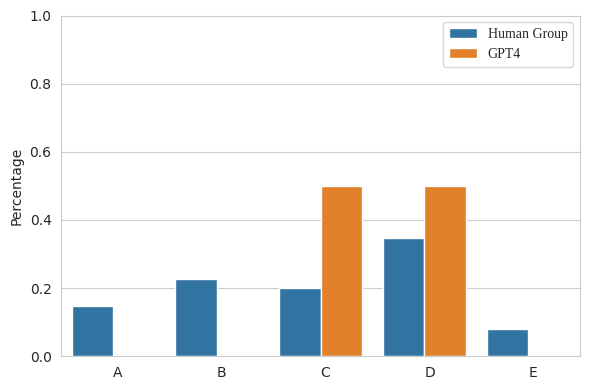}
        \caption{Answer distribution of DS4 Q10.}
        \label{fig:ds4_q10}
    \end{subfigure}
\caption{Results distribution for perception-based questions for DS1, DS2, DS3 and DS4. ``A'' to ``E'' represents \{``A:Extremely easy'', ``B:Somewhat easy'', ``C:Neither easy nor difficult'', ``D:Somewhat difficult'', ``E:Extremely difficult''\}, respectively.}
\vspace{-.1in}
\label{fig:sel}
\end{figure}

\subsection{Misalignment Analysis}

While the LLM demonstrates strong persona-simulation capabilities, significant misalignments still occur in certain cases. In this section, we take \textit{\{Q10: Please rate the difficulty in understanding this discharge instruction\}} as an example to analyze how these misalignments arise. For Q10, respondents could choose from five different answers: A) Extremely easy, B) Somewhat easy, C) Neither easy nor difficult, D) Somewhat difficult, and E) Extremely difficult. The distribution of responses from both the LLM and human participants is shown in \autoref{fig:sel}.

\begin{tcolorbox}[left=1mm, right=1mm, top=0.5mm, bottom=0.5mm, arc=1mm]
\textbf{Finding 5:} The LLM performs better at simulating human groups with more homogeneous response patterns.
\end{tcolorbox}

As seen in \autoref{fig:sel}, the LLM's selection patterns are generally more concentrated, often covering only one or two adjacent response categories. For example, in the case of shorter discharge summaries (DS1), the LLM's predictions are focused on responses C) Neither easy nor difficult, with minimal variation across the other options. In contrast, the distribution of human responses spans all five categories, reflecting the more diverse nature of human comprehension levels. This pattern indicates that the LLM is better at simulating personas within groups where responses are more homogeneous, as it tends to assign probabilities heavily to a smaller set of options. 

In real-world scenarios, however, human responses to complex information, such as discharge instructions, tend to be more dispersed due to varying levels of familiarity, personal experience, and cognitive abilities. Therefore, the LLM's centralized predictions reveal a limitation in capturing the full range of human diversity, especially in scenarios where individuals interpret the same information very differently.

\begin{tcolorbox}[left=1mm, right=1mm, top=0.5mm, bottom=0.5mm, arc=1mm]
\textbf{Finding 6:} The LLM tends to overestimate human comprehension, particularly when processing long discharge summaries.
\end{tcolorbox}

A deeper analysis reveals that the LLM consistently overestimates human comprehension when dealing with longer discharge summaries, such as DS2 and DS3. In these cases, the LLM only selects \textit{A) Extremely easy} or  \textit{B) Somewhat easy} as the response, indicating that it perceives the information to be straightforward. However, human participants often rated these longer summaries as more challenging, with a significant portion of responses falling under D) Somewhat difficult and E) Extremely difficult. For example, in DS3, 30\% of human respondents chose D or E, while the LLM allocated no responses to these categories, reflected in Figure~\ref{fig:ds3_q10}, suggesting a clear overestimation of comprehension difficulty by the LLM.

This discrepancy can likely be attributed to the LLM's inherent ability to process large amounts of text without difficulty. While it can parse lengthy content efficiently, it may not fully account for the cognitive load that such content places on human readers. Lengthier discharge instructions often contain more complex language and medical jargon, which can overwhelm readers, especially those with lower health literacy. The LLM, however, lacks the ability to fully contextualize how factors like summary length, language complexity, and familiarity with medical terminology can affect human understanding, leading to misalignment in its predictions.


\subsection{Case Studies}

\begin{figure}[ht]
    \centering
    \begin{subfigure}[t]{.3\columnwidth}
        \centering
        \includegraphics[width=\textwidth]{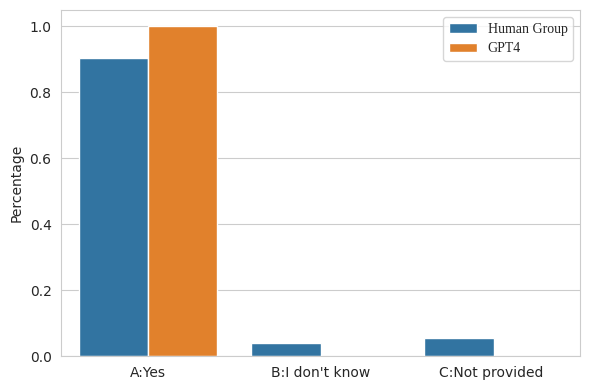}
        \caption{Answers distribution of Q2: ``Do you know the names of all your medications?''}
        \label{fig:ds1_q2}
    \end{subfigure}
    \hspace{0.02\columnwidth}
    \begin{subfigure}[t]{.3\columnwidth}
        \centering
        \includegraphics[width=\textwidth]{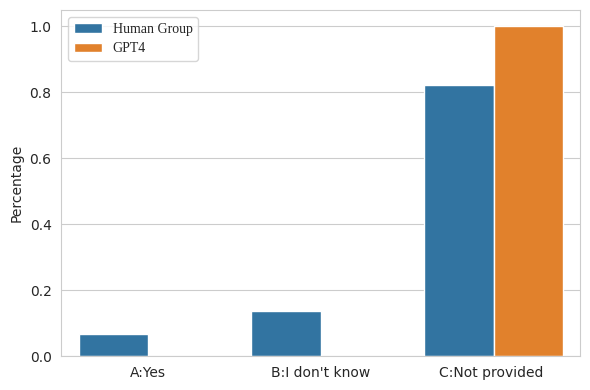}
        \caption{Answers distribution of Q3:``Do you know your diagnosis?''}
        \label{fig:ds2_q3_comp}
    \end{subfigure}
    \hspace{0.02\columnwidth}
    \begin{subfigure}[t]{.3\columnwidth}
        \centering
        \includegraphics[width=\textwidth]{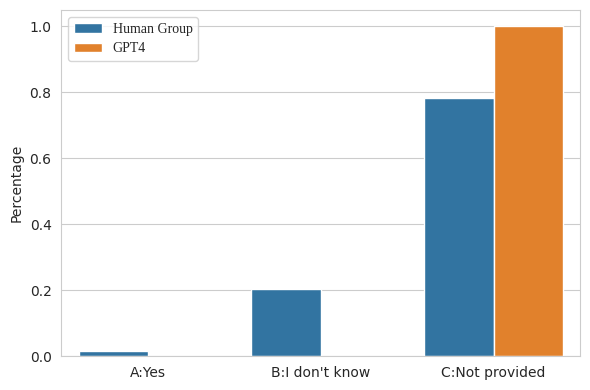}
        \caption{Answers distribution of Q4:``Do you know the common side effects of all your medications?''}
        \label{fig:ds3_q4}
    \end{subfigure}
\caption{Examples of Answers distributions of Selected Questions.}
\vspace{-.1in}
\label{fig:good_exampe}
\end{figure}

In this section, we analyze the questions with the highest and lowest alignment rates to gain deeper insights into the LLM's performance.
For questions with high alignment rates, we focus on \textit{\{Q2: Do you know the names of all your medications?\}}, \textit{\{Q3: Do you know your diagnosis?\}}, and \textit{\{Q4: Do you know the common side effects of all your medications?\}}. The possible answers to these questions are: \textit{A) Yes, B) I don't know, and C) Not provided.}

From \autoref{fig:good_exampe}, we can observe that for all of these questions, both real human responses and the LLM tend to answer A) Yes or C) Not provided. The high alignment rate may be due to the nature of the questions, which involve factual knowledge (e.g., knowing medication names or diagnoses). These types of questions are relatively straightforward, and the LLM can easily match human responses by recognizing whether the information is present. This could be why the LLM excels in this context, as the binary nature of the answers (Yes or Not provided) simplifies the task of aligning with human responses. Additionally, the structured and factual nature of the questions likely makes it easier for the LLM to rely on knowledge retrieval or pattern recognition.

However, it is worth noting that the LLM never selects \textit{B) I don't know}, while real human respondents occasionally do. This discrepancy may reflect a limitation in the LLM's personas capability. Human respondents are more likely to express uncertainty or admit a lack of knowledge, which is a natural human behavior when dealing with personal health information. The LLM, on the other hand, may be overconfident, defaulting to providing answers or skipping the question (selecting ``Not provided'') rather than expressing uncertainty. This could stem from its training, where it is less likely to be exposed to scenarios requiring expressions of doubt or lack of knowledge. Another potential reason is that the LLM may prioritize generating responses that appear complete or assertive, even in cases where human-like uncertainty would be more appropriate.

This tendency could indicate an area where the LLM's persona modeling falls short, particularly when simulating responses that require expressing uncertainty or acknowledging gaps in knowledge—traits that are common in real human behavior.

For questions with low alignment rates, we focus on \textit{\{Q5: Do you know your diagnosis?\}} and \textit{\{Q8: "Are there any activities or foods you need to avoid?\}} For Q5, the possible answers are: \textit{A) Yes, B) I don't know, and C) Not provided.} For Q8, the options are: \textit{A) Avoid fruit, B) Avoid strenuous exercise, C) Other, and D) I don't know.}

\begin{figure}[ht]
    \centering
    \begin{subfigure}[t]{.3\columnwidth}
        \centering
        \includegraphics[width=\textwidth]{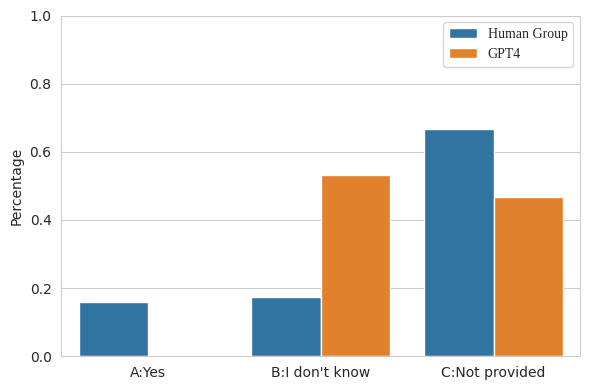}
        \caption{Answers distribution of Q5: ``Do you know your diagnosis?''}
        \label{fig:ds2_q5}
    \end{subfigure}
    \hspace{0.05\columnwidth}
    \begin{subfigure}[t]{.35\columnwidth}
        \centering
        \includegraphics[width=\textwidth]{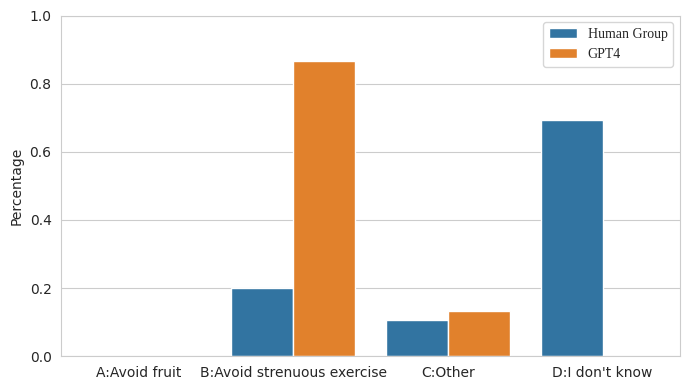}
        \caption{Answers distribution of Q8:``Are there any activities or foods you need to avoid?''}
        \label{fig:ds3_q8}
    \end{subfigure}
\caption{Examples of Answers distributions of Selected Questions.}
\vspace{-.1in}
\label{fig:bad_exampe}
\end{figure}

From \autoref{fig:bad_exampe}, we can observe that for Q5, the LLM tends to select B) I don't know more frequently, whereas real human participants are more likely to choose C) Not provided. One potential reason for this discrepancy could be the LLM's over-reliance on statistical patterns from its training data. Since ``I don't know'' is often a common response in ambiguous or unclear situations, the LLM may interpret this question as one where individuals are uncertain about their diagnosis, defaulting to that response. However, human participants may be more aware of the context in which this question is typically asked. In healthcare settings, when patients don't receive clear communication regarding their diagnosis, they often recognize that the information has not been explicitly provided, rather than expressing uncertainty. Therefore, they are more likely to select ``Not provided''. This difference reflects the challenge LLMs face in capturing subtle contextual cues, such as understanding that the absence of information (rather than uncertainty) might be a more appropriate response in certain scenarios.

For Q8, the LLM demonstrates a preference for B) Avoid strenuous exercise, while human respondents more often select D) I don't know. This result could stem from the LLM's tendency to favor more definitive or action-oriented responses, especially when dealing with health-related queries. In many cases, medical advice includes clear recommendations like avoiding strenuous exercise, which is commonly associated with physical restrictions after treatment or surgery. Consequently, the LLM may lean towards selecting such specific advice, even when the context of the question is broader or less definitive. On the other hand, human respondents are often more cautious in giving such specific answers unless they have been directly instructed by healthcare professionals. In the absence of explicit guidance or when unsure, humans are more likely to select ``I don't know''. This reflects a fundamental difference in how LLMs and humans handle ambiguity: while the LLM attempts to provide a plausible answer based on general knowledge, humans may defer to their personal experience, recognizing uncertainty and the need for more personalized medical advice. This highlights a limitation in the LLM’s ability to accurately simulate the uncertainty and hesitation that humans naturally exhibit in complex medical scenarios.

%% file: demo_tables.tex
\begin{table}[h]

\footnotesize
\caption{Demographics of participants.}
\begin{tabular}{clll}
\toprule
\textbf{Item}                                          & \textbf{Options}          & \textbf{n} &  \\ \midrule
\multirow{3}{*}{\textbf{Gender}}                       & Male  & 58          &       \\
                                                       & Female  & 35          &          \\
                                                       & Non-binary  & 6           &            \\
                                                       \midrule
\multirow{5}{*}{\textbf{Age}}              & 18-24  &  21        &             \\
                                           & 25-34   &  38        &             \\
                                           & 35-44   &   23       &             \\
                                           & 45-54   &   11       &             \\
                                           & 55-64   &   5       &             \\ \midrule
\multirow{5}{*}{\textbf{Ethnicity}}          & White/Caucasian  &  65     &             \\
                                             & Black/African  &    14        & \\
                                             & Hispanic/Latino  &  5         &             \\
                                             & Asian &    3       &   \\ 
                                             & Mixed Race &    7       &   \\                                              \midrule
\multirow{6}{*}{\textbf{Education}}   
                                                      & No High School  &  3  &      \\
                                                      & Some High School  &  5  &      \\
                                                      & High school graduate  &  27  &      \\
                                                       & Some college    &  12   &        \\
                                                       & College graduate  &    39  &       \\
                                                       & Post-graduate degree   &    11        &  
                                                                \\ \midrule
\multirow{4}{*}{\textbf{Doctor Visit}}   
                                                      & Weekly  &  2  &      \\
                                                      & Monthly  &  21  &      \\
                                                      & Yearly  &  71  &      \\
                                                       & Never    &  4   &      
                                                                \\ \midrule
\multirow{4}{*}{\textbf{ER Visit}}   
                                                      & Weekly  &  1  &      \\
                                                      & Monthly  &  2  &      \\
                                                      & Yearly  &  51  &      \\
\bottomrule
\end{tabular}
\label{tab:demographics}
\end{table}

%% file: 5.discussion.tex
\section{Broader Impact}



In this paper, we focus on evaluating the role-playing capabilities of LLMs, examining how well they can mimic individuals from varied demographic and experiential profiles. Using discharge instructions as a test case, we explore whether the responses generated by LLMs align with the understanding and behavior of actual patients. Through this study, we highlight the potential of LLMs to replace human feedback in training phases, such as in RLHF, and demonstrate their capacity to simulate human-like responses, ultimately contributing to the development of more inclusive and effective patient communications. These insights serve as a first step towards leveraging LLMs to automate healthcare systems while ensuring that communications remain patient-centric and tailored to diverse needs.

We believe that a better understanding of the role-playing capabilities in LLMs can not only advance automated, patient-tailored healthcare communication but also reduce the dependency on traditional, resource-intensive user studies. A strong role-playing capability in LLMs could allow these models to be supplementary to real human feedback in key processes, such as Reinforcement Learning from Human Feedback (RLHF), where human users are typically required to guide models toward more accurate and human-like behavior. For example, if LLMs can effectively impersonate diverse patient backgrounds, they could provide simulated feedback that reflects real-world patient interactions, enabling the system to fine-tune responses without direct human intervention.

However, we have also identified failure modes of LMM for exactly the categories of discharge information where automated assistance may be most valuable. Lower educated and female patients are less well represented by LMMs. This may reflect historical data from medical research that excludes women~\cite{baylis1996women, mazure2015twenty}. Significant gender biases in LLM-generated content have been analyzed by Wang et al. in ~\cite{wan2023kelly}.

We are targeting specially adjusted and trained medical LLMs based on the limited data collected from minority populations. One of the limitations for current LLMs replacing human participants is they can misportray and flatten identity groups as noted by~\cite{wang2024large,bender2021dangers}. ``Misportray'' represents the case where LLMs prompted with a demographic identity will more likely represent what \textbf{out-group members} think of that group than what \textbf{in-group members think of themselves}. A utilized LLM role-playing should learn and understand from these specific groups, not using a ``traditional'' and stereotype impersonation.   Direct adoption of LLMs without evaluating their efficacy in diverse heterogeneous populations risks creating a situation in which only highly educated male participants are correctly represented. This has historically been a challenge as medical studies have focused on well-educated male participants, resulting in high levels of bias in existing data~\cite{geller2011inclusion}.

LLMs responses to the same prompt are inherently identical or leptokurtic. LLMs are well-suited to support populations that are self-similar, such as individuals with similar levels of formal education in a given field. In addition, it is easier to model the answers of individuals who understand factual instructions than the answers of those who are confused. Additional qualitative research is needed to better understand how low-education individuals approach and process discharge instructions. 



This could greatly enhance the clarity of medical communications like discharge instructions, ensuring that the information is more comprehensible and accessible to all patients with different medical knowledge levels, languages, education levels, races, or cultural backgrounds.

%% file: 6.relatedwork.tex
\section{Related Work}

In-context learning allows large language models (LLMs) to adapt to tasks by leveraging context provided in input prompts~\cite{brown2020language}. This approach differs from conventional methods that require explicit model fine-tuning, as LLMs can instead use contextual information without modifying their underlying parameters~\cite{lampinen2022can,wei2022chain,arora2022ask}. Prompts, typically in textual form, provide scenarios or questions, guiding LLMs to generate relevant responses~\cite{zhou2022large}. 

One strength of in-context learning is its flexibility across different learning regimes. LLMs can perform effectively even when no examples are given, demonstrating strong zero-shot capabilities~\cite{xian2018zero}. Success in areas like reasoning and problem-solving, even without example-based prompts, highlights their adaptability~\cite{yuan2023well,kiciman2023causal}. 

Prompt engineering is a vital component of this process, involving both manual and automated approaches to crafting the most effective prompts~\cite{reynolds2021prompt,sshin2020autoprompt}. The development of optimal prompt strategies has become a specialized area of study, as the design of effective prompts can significantly impact performance~\cite{oppenlaender2023prompting}.

A notable technique is the use of persona-based prompts, where LLMs are guided to adopt specific roles or perspectives~\cite{han2022meet,keskar2019ctrl}. This technique allows the model to generate outputs as if it were a particular character, which can influence the tone, style, or accuracy of responses~\cite{Shanahan2023RolePlayWL}. In some cases, this approach has been used to elicit more truthful responses by framing questions within a particular persona~\cite{lin2021truthfulqa}, while in others, adversarial prompts can push the model towards undesirable behaviors~\cite{wolf2023fundamental}. 

The adaptability of LLMs extends beyond just language-based applications. They are capable of exhibiting behaviors influenced by the context provided, and researchers have used them to simulate various types of interactions~\cite{aher2022using}. Additionally, LLMs can reflect different personalities depending on how the prompts are structured, with their responses shifting to reflect emotional or other human-like traits~\cite{pellert2023ai,coda2023inducing}. This flexibility has enabled their use in simulating complex human behaviors and conditions, including generating personalized responses based on demographic information~\cite{argyle2022out}.

Despite their strengths, LLMs are susceptible to inheriting biases from the data they are trained on. These biases can manifest in their outputs, and under certain conditions, such as when persona-based prompts are used, they can be magnified~\cite{caliskan2017semantics,bender2021dangers,abid2021persistent}. This presents challenges in managing fairness and neutrality in model responses~\cite{coda2023inducing,deshpande2023toxicity}.

Finally, LLMs are frequently integrated with models that handle both language and vision tasks. These vision-language models have expanded LLMs' capabilities, particularly in tasks such as image classification and visual question answering~\cite{radford2021clip,menon2023visual,Yang2022LanguageIA}. By synthesizing language and visual data, these combined models offer new opportunities for improving performance in multi-modal tasks.

%% file: 7.conclusion.tex
\section{Conclusion and Future Works}

This study demonstrates the potential of Large Language Models (LLMs) to simulate diverse individuals in healthcare settings, specifically for the comprehension of discharge summaries. By employing role-playing prompts, LLMs can adapt to personas with various backgrounds, providing an innovative method to customize complex medical information for different audiences. Our findings suggest that LLMs, when guided by structured prompts, can generate responses that align with the understanding of individuals from diverse backgrounds, demonstrating their promise in enhancing patient communication. However, the role-playing capabilities of LLMs are not without limitations. Over-simplification and the nuances of individual comprehension present challenges that must be addressed to fully realize their potential. These limitations indicate that while LLMs can be a powerful tool for generating personalized content, further refinement is needed to ensure their outputs meet real-world patient needs effectively.

Future work will focus on enhancing the accuracy and consistency of LLM role-playing, ensuring that these models can better capture subtle variations in individual comprehension. Additionally, research should explore more sophisticated techniques for integrating LLMs into healthcare systems, such as incorporating real-time patient feedback to dynamically refine communication strategies and ensure responses are both effective and sensitive to individual needs. 
By continuously improving the role-playing abilities of LLMs and addressing their current limitations, we can move closer to a future where automated, patient-specific healthcare communication becomes both feasible and effective. This progress will not only streamline tasks like discharge summary generation but also elevate the overall patient experience, promoting more inclusive and accessible healthcare for diverse populations. 

%% file: appendix.tex
\appendix
\section*{Appendix}

\section{Prompt Example with Lower Education Level Persona}
\label{apdx:propt}
You are someone who has never received a college degree. You will be reading a Discharge Summary written by a clinician for a patient. After reading the Discharge Summary, I will ask you a question. Your job is to think of someone with your background and choose the correct answer by selecting the letter of the answer (e.g., A, B, C, etc.). You should only output the letter with your final answer without any explanation.

Here is the Discharge Summary:
“You were admitted and found to have an ulcer in the duodenum. To help this heal, we are prescribing new medications (pantoprazole). Please be sure to take this until you are seen in follow-up.”

Now, answer the following question:
Question:
Please rate your understanding level of this discharge instruction.
A. Very clear
B. Somewhat clear
C. Not clear at all
Your task: Only respond with the letter of the answer that best reflects your understanding.

\section{Discharge Summary Samples}
\label{apdx:discharge}

\begin{enumerate}
    \item[DS1] You were admitted and found to have an ulcer in the duodenum. To help this heal, we are prescribing new medications (pantoprazole). Please be sure to take this until you are seen in follow-up.

    \item[DS2] Call Dr. xxx if experience: \\
    -Take stool softeners with narcotics \\
    -Fever > 101 or chills \\
    -Increased shortness of breath or cough \\
    -Chest pain \\
    -You may shower. No swimming for 4 weeks \\
    -No driving while taking narcotics.

    \item[DS3] Keep splint/dressing on until follow-up \\
    Keep splint clean \& dry at all times \\
    Follow up with Dr. xxx in xxx days \\
    Wean off of narcotics \\
    Take aspirin for 2 weeks \\
    Physical Therapy: \\
    NWB left lower extremity \\
    Leave splint on until follow-up

    \item[DS4] Ok to shower today but wear tegaderm dressing over the drain site.  No heavy lifting. Return to ED for anything that concerns you.
\end{enumerate}

\section{Survey Questions for Each Discharge Summary}
\label{apdx:survey_questions}
\subsection{Information-based Questions}
\begin{enumerate}
    \item[Q2] Do you know the name of all your medications? (If yes, please type.) A. Yes B. I don't know C. Not provided

    \item[Q3] Do you know your diagnosis? (If yes, please type.) A. Yes B. I don't know C. Not provided

    \item[Q4] Do you know the common side effects of all your medications? (If yes, please type.) A. Yes B. I don't know C. Not provided

    \item[Q5] Are there other prescriptions given besides the medication? (If yes, please type.) A. Yes B. I don't know C. Not provided

    
    \item[Q6] 


    Do you know what kind of condition you have mentioned in the discharge instructions? A. Stomach disease  B. Ulcer in the duodenum   C. Wearing Tegaderm C. Keep splint D. I don't know   

    \item[Q7] Do you know what kind of treatment you need to follow based on the discharge instructions? A. Nothing B. Take a new medication C.  See a doctor again D.  I don't know

    \item[Q8] Are there any activities or foods you need to avoid? A. Avoid fruit B. Avoid strenuous exercise C.  Others D. I don't know

    \item[Q9] Is there anything about your discharge instructions that is unclear or worrying you? A. Medication schedule B. Follow-up appointments C. Symptoms to watch for D. Dietary restrictions E. Activity limitations F. Other, please specify. G. No, it's very clear
\end{enumerate}

\subsection{Perception-based Questions}
\label{apdx:prcp_q}
\begin{enumerate}
    \item[Q1] Please rate your understanding level of this discharge instruction. A. Very clear. B. Somewhat clear. C. Not clear at all.

    \item[Q10] Please rate the difficulty in understanding this discharge instruction. A. Extremely easy B. Somewhat easy C. Neither easy nor difficult D. Somewhat difficult E.  Extremely difficult
\end{enumerate}